\documentclass[letterpaper]{article} % DO NOT CHANGE THIS
\usepackage{aaai2026}  % DO NOT CHANGE THIS
\usepackage{times}  % DO NOT CHANGE THIS
\usepackage{helvet}  % DO NOT CHANGE THIS
\usepackage{courier}  % DO NOT CHANGE THIS
\usepackage[hyphens]{url}  % DO NOT CHANGE THIS
\usepackage{graphicx} % DO NOT CHANGE THIS
\usepackage{natbib}  % DO NOT CHANGE THIS AND DO NOT ADD ANY OPTIONS TO IT
\usepackage{caption} % DO NOT CHANGE THIS AND DO NOT ADD ANY OPTIONS TO IT
\usepackage{algorithm}
\usepackage{algorithmic}

\usepackage{amsmath} 

\usepackage{newfloat}
\usepackage{listings}
\DeclareCaptionStyle{ruled}{labelfont=normalfont,labelsep=colon,strut=off} % DO NOT CHANGE THIS
\floatstyle{ruled}
\newfloat{listing}{tb}{lst}{}
\floatname{listing}{Listing}
\title{ProFuser: Progressive Fusion of Large Language Models}
\author{
    Tianyuan Shi\textsuperscript{\rm 1}, Fanqi Wan\textsuperscript{\rm 1}, Canbin Huang\textsuperscript{\rm 1}, Xiaojun Quan\textsuperscript{\rm 1}\thanks{$\;\;$Corresponding authors.}, \\ Chenliang Li\textsuperscript{\rm 2}, Ming Yan\textsuperscript{\rm 2}, Ji Zhang\textsuperscript{\rm 2}, Minhua Huang\textsuperscript{\rm 3}, Wu Kai\textsuperscript{\rm 3}
}
\affiliations{
    \textsuperscript{\rm 1}School of Computer Science and Engineering, Sun Yat-sen University\\
    \textsuperscript{\rm 2}Alibaba Group, \textsuperscript{\rm 3}China Mobile Internet\\
    \{shity6, wanfq, huangcb3\}@mail2.sysu.edu.cn, quanxj3@mail.sysu.edu.cn \\
    \{lcl193798, ym119608\}@alibaba-inc.com, huangminhua@cmic.chinamobile.com
}
         
\usepackage{bibentry}
\begin{document}

\maketitle

\begin{abstract}
While fusing the capacities and advantages of various large language models offers a pathway to construct more powerful and versatile models, a fundamental challenge is to properly select advantageous model during training. 
Existing fusion methods primarily focus on the training mode that uses cross entropy on ground truth in a teacher-forcing setup to measure a model's advantage, which may provide limited insight towards model advantage. In this paper, we introduce a novel approach that enhances the fusion process by incorporating both the training and inference modes. Our method evaluates model advantage not only through cross entropy during training but also by considering inference outputs, providing a more comprehensive assessment. To combine the two modes effectively, we introduce ProFuser to progressively transition from inference mode to training mode. To validate ProFuser's effectiveness, we fused three models, including Vicuna-7B-v1.5, Llama-2-7B-Chat, and MPT-7B-8K-Chat, and demonstrated the improved performance in knowledge, reasoning, and safety compared to baseline methods.
\end{abstract}

\begin{links}
    \link{Code}{https://github.com/Stycoo/ProFuser}
    % \link{Datasets}{https://aaai.org/example/datasets}
    % \link{Extended version}{https://aaai.org/example/extended-version}
\end{links}

\section{Introduction}

Large Language Models (LLMs) have demonstrated remarkable capabilities across diverse tasks in recent years. However, their training demands substantial computational resources, often requiring thousands of GPUs and processing trillions of tokens \cite{sukhbaatar2024branchtrainmix}. In this context, integrating the complementary capabilities of existing LLMs into a unified model presents a resource-efficient alternative for achieving enhanced performance.

Traditional approaches to capability integration often leverage ensemble methods \cite{monteith2011turning,llmblender}, which combine predictions from multiple trained models during inference. While effective, these methods necessitate the simultaneous deployment of multiple models, introducing significant memory and computational overhead, particularly challenging for resource-intensive LLMs.
An alternative paradigm focuses on parameter-space merging, where multiple models are consolidated through arithmetic operations on their parameters \cite{gupta2020}. This approach requires determining optimal combination coefficients, either through manual calibration \cite{wortsman2022model,yadav2024ties} or automated optimization \cite{Matena2021,jin2023}. However, these methods are fundamentally constrained by the requirement for architectural homogeneity among models.
To address these limitations, FuseLLM \cite{wan2024knowledge} introduces a pioneering approach enabling fusion of architecturally heterogeneous LLMs. Grounded in knowledge distillation, FuseLLM transfers collective knowledge from multiple source LLMs to a target LLM through probability distribution matrices.

\begin{figure*}[htbp]
    \centering
    \includegraphics[width=\linewidth]{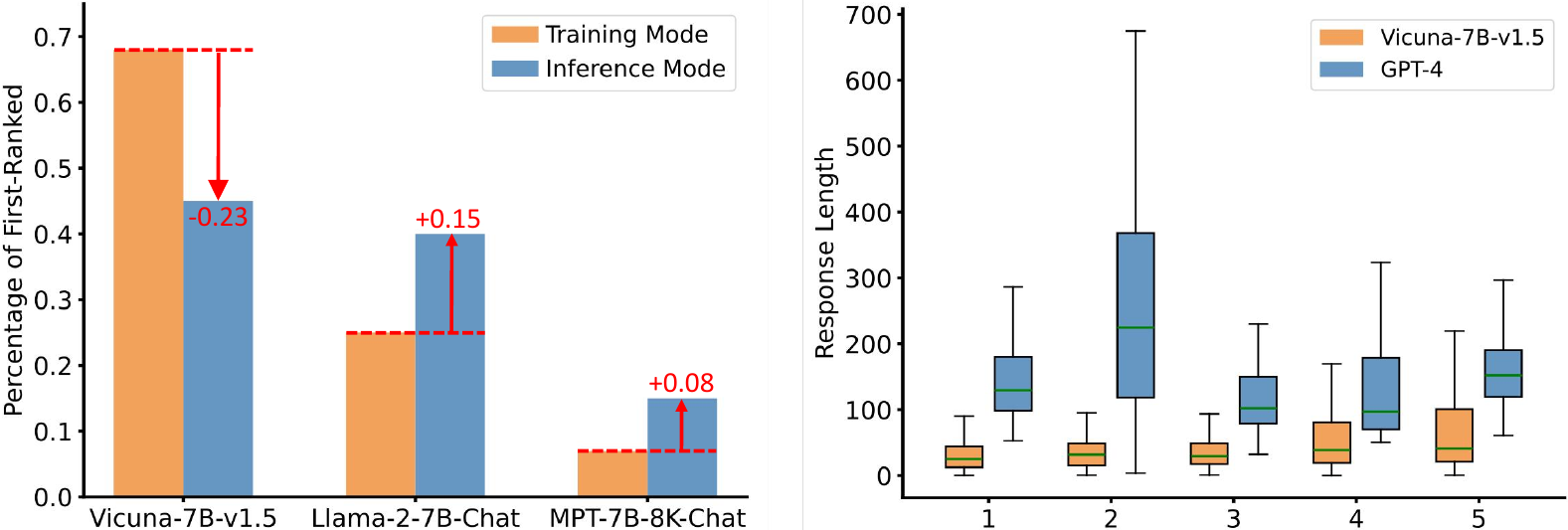}
    \caption{\textbf{Left}: Performance comparison between Vicuna-7B-v1.5 (Vicuna) and other source models across training and inference modes. In training mode, Vicuna outperforms Llama-2-7B-Chat on 68\% of evaluation samples (measured by Min-CE), highlighting its superior token prediction capability. However, this advantage diminishes in inference mode, where Vicuna's success rate drops to 45\% (evaluated using reward models). This highlights a gap between token-level prediction and response generation quality. \textbf{Right}: Response length comparison between GPT-4 and Vicuna-7B-v1.5 for the five most frequently occurring system messages in the training set (x-axis IDs correspond to specific system prompts). GPT-4 consistently produces longer and more detailed responses across all prompt types.} 
    \label{preliminary_exp}
\end{figure*}

While FuseLLM demonstrates promising results, its reliance on minimum cross-entropy (Min-CE) in teacher-forcing training mode for assessing source model capabilities may not fully capture their strengths in real-world inference scenarios. 
Our empirical analysis, illustrated in Figure \ref{preliminary_exp} (left), evaluates models in both training and inference modes to quantify this limitation. Training mode assessment employs Min-CE to measure ground-truth (GT) token prediction accuracy, while inference mode utilizes reward models to evaluate generated response quality.
Our investigation reveals a significant performance disparity: while Vicuna-7B-v1.5 demonstrates superiority over Llama-2-7B-Chat in 68\% of training mode cases, this advantage diminishes to 45\% in inference mode, achieving parity with Llama-2-7B-Chat. This discrepancy emerges from the fundamental difference between next-token prediction proficiency in training mode and response generation quality in inference mode.

Effective model fusion necessitates comprehensive advantage exploitation across both training and inference modes. However, our experimental results (Section \ref{main_results}) indicate that simultaneous optimization across both modes yields suboptimal improvements when inference mode is weighted minimally. This phenomenon can be attributed to the qualitative distinction between the "advantage carriers" in each mode: training mode utilizes more complex and detailed GT outputs compared to the relatively concise source model outputs in inference mode (Figure \ref{preliminary_exp} (right)). 
To address this challenge and answer the question: \textit{How can we effectively leverage advantages from both modes for optimal fusion?} we propose \textbf{ProFuser}, inspired by progressive learning principles \cite{mukherjee2023orca}. ProFuser implements a two-stage fusion strategy: (1) \textit{Inference mode fusion}, which prioritizes high-quality response generation capabilities, is followed by (2) \textit{Training mode fusion}, which incorporates next-token prediction strengths. This progressive approach facilitates effective integration by bridging the qualitative gap between modes.

We validate ProFuser's effectiveness by integrating capabilities from Vicuna-7B-v1.5, Llama-2-7B-Chat, and MPT-7B-8K-Chat into Vicuna-7B-v1.5-ProFuser. Experimental results demonstrate consistent improvements across knowledge, reasoning, and safety dimensions. Further analysis validates our dual-mode advantage evaluation framework's ability to identify model strengths consistently, even with relatively weaker source models (e.g., MPT-7B-8K-Chat). Moreover, the progressive fusion strategy exhibits enhanced learning stability.

\begin{figure*}[htbp]
    \centering
    \includegraphics[scale=0.8]{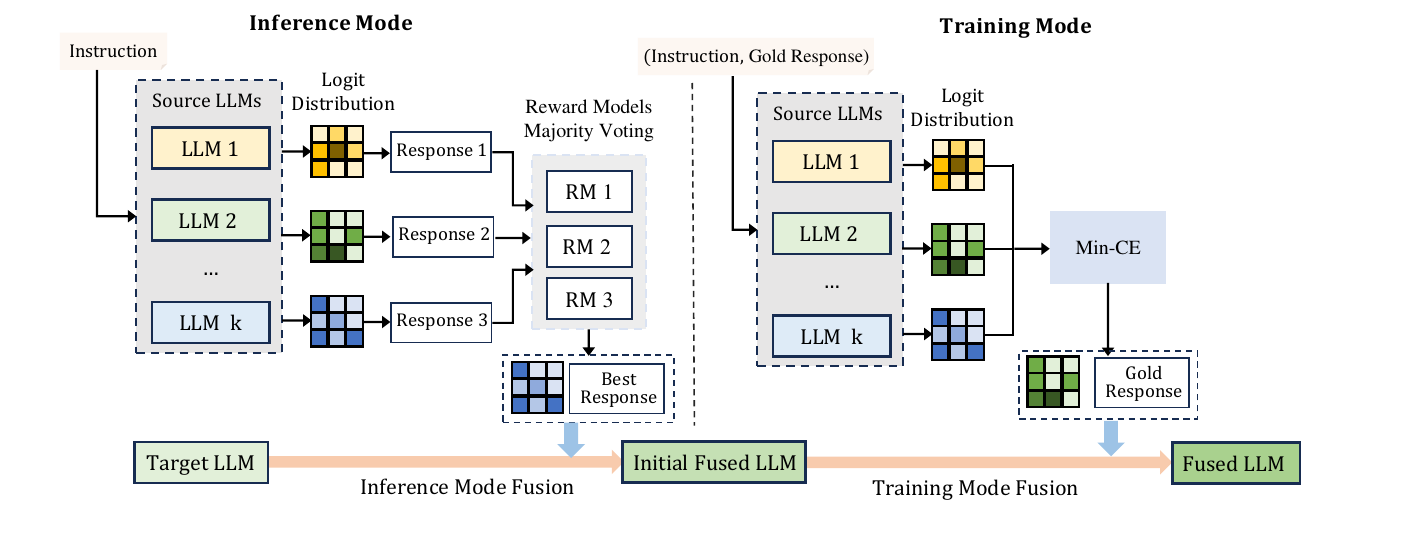}
    \caption{Overview of the Progressive Model Fusion Method (ProFuser). The framework operates in two sequential stages: inference mode and training mode. In inference mode, reward models (RM) evaluate response quality to identify advantageous outputs, while in training mode, minimum cross-entropy (Min-CE) determines optimal token distributions. Heterogeneous source LLMs (represented by distinct colors) contribute their respective advantages, which are progressively integrated into the target model through an easy-to-hard learning paradigm. This dual-mode approach ensures comprehensive capability transfer from source models to the target model.}
    \label{method}
    \vspace{-5mm}
\end{figure*}
% \vspace{-0.15cm}
\section{Related Work}
% \subsection{Knowledge Distillation}
\paragraph{Knowledge distillation} (KD, \citet{hinton2015distilling}) represents a systematic approach to compress knowledge from one or more large teacher models into a more compact student model while maintaining performance efficacy. In text classification domains, various methodologies have been proposed: output distribution mimicking \cite{turc2019,zhang2023}, hidden state replication \cite{sun2019patientknowledgedistill,jiao2020tinybert}, and attention score emulation \cite{wang2021minilmv2}. For text generation tasks, knowledge transfer strategies include learning from teacher logits distributions on ground truth sequences \cite{agarwal2024,gu2024minillm} or generated outputs \cite{peng2023instructiontuninggpt4}. Multi-teacher knowledge distillation (MTKD) enhances distillation effectiveness through distribution averaging \cite{you2017learning} or sequence blending \cite{wang2024openchat} across multiple teachers.
While KD focuses on model compression, model fusion pursues a fundamentally different objective: synthesizing complementary capabilities from multiple source models to create a more comprehensive and capable unified system.
\vspace{-0.1cm}
\paragraph{Model merging} encompasses techniques for combining weights from multiple models through direct parameter space manipulation. Contemporary approaches can be categorized into two primary paradigms:
(1) \textit{Same-task model merging} primarily focuses on enhancing generalization capabilities. For instance, Model Soups \cite{wortsman2022model} implements linear averaging of models fine-tuned through diverse strategies on identical datasets. However, these approaches are constrained by the requirement for homogeneous training protocols and typically yield incremental improvements.
(2) \textit{Cross-task model merging} aims to facilitate multi-task learning (MTL). Fisher Merging \cite{Matena2021} employs Fisher information matrices for parameter weighting, though its computational complexity becomes prohibitive for large-scale models. RegMean \cite{jin2023} reformulates merging as an optimization problem, minimizing the L2 distance between the merged model and individual constituents. Task Arithmetic \cite{zhang2023} introduces "task vectors" for parameter-efficient merging via LoRA \cite{hu2021lora}, while PEM Composition \cite{zhang2023} extends this framework to LoRA-based architectures. Ties-Merging \cite{yadav2024ties} addresses task conflicts through systematic parameter pruning and sign alignment.

A significant limitation of existing approaches is their requirement for architectural homogeneity among models. FuseLLM \cite{wan2024knowledge} addresses this constraint by enabling fusion of heterogeneous LLMs through knowledge distillation, utilizing Min-CE on ground truth to identify advantageous models and transferring their logits distributions to a target LLM. Our work extends this framework by introducing a comprehensive advantage evaluation mechanism that considers both training and inference modes, thereby enabling a more thorough assessment and integration of model capabilities.

\section{Preliminaries}
Given an instruction dataset \( D = \{(x_i, y_i)\}_{i=1}^N \), where \( x_i \) and \( y_i \) denote the \( i \)-th instruction and its corresponding response, respectively. Supervised Fine-Tuning (SFT) aims to refine pre-trained language models parameterized by \( \theta \) to develop instruction-following capabilities through supervised learning. This is achieved by minimizing the log-likelihood loss:  
\begin{equation}
    \mathcal{L}_{\text{SFT}}(x_i, y_i) = -\sum_{t=1}^T \log p_{\theta}(y_{i,t} \mid x_i, y_{i,<t}),
    \label{sft-loss}
\end{equation}  
where \( T \) is the length of response \( y_i \), \( p_{\theta}(y_{i,t} \mid x_i, y_{i,<t}) \) represents the probability of predicting the \( t \)-th ground-truth (GT) token \( y_{i,t} \), conditioned on the instruction \( x_i \) and all preceding GT tokens \( y_{i,<t} \).  
Teacher-forcing is used during training, where the model receives the ground-truth tokens \( y_{i,<t} \) as input rather than its own predictions.  

\section{Method}
Our research focuses on integrating advantageous capabilities from multiple source models into a target model through model fusion. This process presents two fundamental challenges:
1. \textit{Advantage evaluation}: Previous approaches rely solely on ground truth Min-CE (training mode) for advantage assessment, providing limited insights. We propose a comprehensive evaluation framework incorporating both inference and training modes, enabling a more thorough identification of model strengths and facilitating more effective fusion.
2. \textit{Fusion strategy}: Leveraging the differential characteristics of multi-modal advantage information, we introduce a progressive fusion strategy that implements an easy-to-hard learning paradigm, transitioning from inference mode to training mode optimization.

\subsection{Dual Mode Advantage Evaluation}
We propose a dual-mode evaluation framework encompassing both training and inference modes to comprehensively assess model advantages. In training mode, we hypothesize that the probability distribution generated under teacher-forcing for a given \text{(instruction, response)} pair reflects the model's inherent understanding, with lower cross-entropy (CE) values indicating superior performance. In inference mode, response quality serves as a direct indicator of problem-solving capabilities, with higher-quality responses signifying model superiority.

\paragraph{Training Mode} As illustrated in the right side of Figure \ref{method}, given an input instruction \( x_i \) and ground truth response \( y_i \), we employ teacher-forcing to obtain logits distributions \(\{P_i^j\}_{j=1}^K\) from source models \(\{M_j\}_{j=1}^K\). The cross-entropy (CE) is computed for each model following Equation~\eqref{sft-loss}. The advantageous model is identified through minimum CE selection: 
\begin{equation}
     M^{\text{MinCE}} = \text{argmin} (\{L_{\text{SFT}}^{\theta_j}(x_i, y_i)\}_{j=1}^K),
\end{equation}
where $\theta_j$ denotes the parameters of the $j$th source model.
The logits distribution \( P_i^{\text{MinCE}} \) from the selected model encapsulates the training mode advantage information.

\paragraph{Inference Mode} As depicted in the left side of Figure \ref{method}, for instruction \(x_i\), we generate inference outputs \(\{\widetilde{y}_i^j\}_{j=1}^K\) from source models \(\{M_j\}_{j=1}^K\). 
Response quality assessment is conducted through a voting mechanism incorporating multiple high-performance reward models (refer to Section \ref{sec:experiment}, Training Details). The response receiving the majority vote is selected:
\begin{equation}
    \widetilde{y}_i^{\text{B}} = \text{argmax} (\text{RM}_\text{Vote}(\{\widetilde{y}_i^j\}_{j=1}^K)).
\end{equation}
The selected \( \widetilde{y}_i^{\text{B}} \) and its corresponding logits distribution \( \widetilde{P}_i^{\text{B}} \) constitute the inference mode advantage information.

\subsection{Progressive Fusion}
Our progressive fusion strategy capitalizes on the inherent complexity differential between inference mode source model outputs and training mode GPT-4 outputs, with the latter typically exhibiting greater detail and complexity. We implement an easy-to-hard fusion paradigm, sequentially incorporating inference mode followed by training mode.

The capability transfer from source LLMs to the target LLM is achieved through a combination of sequence-level loss $L_{SFT}$ and token-level loss $D_{KL}$:
\begin{equation}
    L_{\text{Fuse}}(x, y, P_S) = L_{\text{SFT}}(x, y) + \beta D_\text{KL}(P_{S}, P_{T}),
    \label{fusion-obj}
\end{equation}
where \(P_S\) and \(P_T\) denote the logits distributions of the advantageous source model and target model with respect to \(y\), respectively.

For a given instruction $x_i$, we incorporate both inference mode advantage information \(\left(\widetilde{y}_i^{\text{B}}, \widetilde{P}_i^{\text{B}}\right)\) and training mode advantage information \(\left(y_i, P_i^{\text{MinCE}}\right)\) into Equation~\eqref{fusion-obj}, yielding mode-specific fusion objectives: \(L_{\text{Infer-Fuse}}(x_i, \widetilde{y}_i, \widetilde{P}_i^{\text{B}})\) and \(L_{\text{Train-Fuse}}(x_i, y_i, P_i^{\text{MinCE}})\).

The comprehensive progressive fusion objective is formulated as:
\begin{equation}
    L_{\text{ProFuser}} = w_1 L_{\text{Infer-Fuse}} + w_2 L_{\text{Train-Fuse}},
\end{equation}
where weights \(w_1\) and \(w_2\) are dynamically adjusted throughout the fusion process. Initially, \(w_1 = 1\) and \(w_2 = 0\) prioritize inference mode advantages. Subsequently, \(w_2\) increases to 1 while \(w_1\) reduces to 0.1, maintaining inference mode insights while emphasizing training mode optimization.
This staged approach ensuring comprehensive capability enhancement in the target LLM.

\begin{table*}[ht]
\centering
\resizebox{0.9\textwidth}{!}{%
\begin{tabular}{lccccccc} \hline
                       & \textbf{MMLU}  & \textbf{HellaSwag} & \textbf{ARC}   & \textbf{Winogrande} & \textbf{GSM8K} & \textbf{TruthfulQA} & \textbf{Average} \\ \hline
MPT-7B-8K-Chat            & 41.55      & 77.52          & 46.93      & 71.35           & 11.00  & 43.70           & 48.68        \\ 
Llama-2-7B-Chat        & 46.74 & \textbf{78.63}          & 52.90      & 71.74           & 16.40 & 44.59           &  51.83       \\
Vicuna-7B-v1.5         & 51.17 & 77.36     & 53.75 & 72.30      & 15.80 & 50.37      & 53.46   \\ \hline
\multicolumn{8}{c}{\textit{Model Fusion}}                                                                 \\ \hline
Vicuna-7B-v1.5-CSFT    & 51.23 & 76.91     & 55.29 & \textbf{74.59}      & 16.76 & 50.39      & 54.20   \\
Vicuna-7B-v1.5-Fuse    & 51.48 & 77.83     & 54.61 & 73.72      & \textbf{18.80} & 50.72      & 54.53   \\
Vicuna-7B-v1.5-ReverseFuse    & 51.09  & 77.87  & 54.69 & 74.19  & 17.21 & 50.77  & 54.30   \\
Vicuna-7B-v1.5-SimulFuse    & 51.54  & 77.74     & 54.95 & 73.64      & 18.77 & 50.74      & 54.56   \\ \hline
Vicuna-7B-v1.5-ProFuser & \textbf{51.85} & 78.39     & \textbf{55.46} & 74.43      & 18.70 & \textbf{51.85}      & \textbf{55.11}  \\ \hline
\end{tabular}%
}
\caption{Overall results of our proposed ProFuser compared against various baseline methods across six benchmarks. Text in \textbf{bold} indicates the best performance. For a detailed explanation of the baseline methods, please refer to Section Baselines.}
\label{tab:main-results}
\end{table*}

\section{Experiments}
\label{sec:experiment}

\subsection{Experimental Setup}
\subsubsection{Source Models and Training Dataset}

Our study draws on three well-established open-source LLM families—Vicuna, Llama, and MPT. Specifically, we employ Vicuna-7B-v1.5~\cite{zheng2023judging}, Llama-2-7B-Chat~\cite{touvron2023llama2}, and MPT-7B-8K-Chat~\cite{MosaicML2023Introducing}.  
Vicuna-7B-v1.5 is designated as the \emph{target} because of its balanced performance and broad task adaptability. To reconcile the heterogeneous tokenizers and vocabularies of these models, we perform token alignment prior to fusion, following~\cite{wan2024knowledge}.
High-quality data are essential for stable fusion. We therefore adopt Orca-Best\footnote{\url{https://huggingface.co/datasets/shahules786/orca-best}}, a semantically deduplicated and quality-filtered subset of the OpenOrca GPT-4 1M instruction corpus~\cite{mukherjee2023orca}. From this collection, we randomly sample 100\,k samples for training.
\vspace{-0.1cm}
\subsubsection{Training Details}
\label{training_details}
% Training is conducted using HuggingFace Transformers~\cite{wolf-etal-2020-transformers} with the Adam optimizer~\cite{kingma2014adam}, a learning rate of \(1.5\times10^{-5}\), cosine annealing decay, a batch size of 128, and a maximum sequence length of 2048.  
% To capture logit distributions from source models on GPT-4 references, top-p=0.95, top-k=10, and temperature=2 are set for both training and inference phases. During inference, one hypothesis per model is sampled. For quality evaluation, three high-performing reward models are selected from RewardBench \footnote{\url{https://huggingface.co/spaces/allenai/reward-bench}}: Eurus-RM-7B~\cite{eurus}, FsfairX-LLaMA3-RM-v0.1~\cite{dong2023raft,xiong2024iterative}, and Starling-RM-7B-alpha~\cite{starling2023}. Predictions are voted upon by these reward models; ties are resolved using the score from the strongest reward model.
% The training process consists of two phases: inference mode fusion and inference-training mode co-fusion. In the first phase, we train for one epoch with the KL loss weight \(\lambda=0.1\). In the second phase, we train for two epochs, with the KL loss weights \(\lambda\) and \(\beta\) set to 0.5, and the mode loss weights \(w_1\) and \(w_2\) set to 0.1 and 1, respectively. The procedure consumes approximately 96 A100-80G GPU-hours.

Our model is trained with HuggingFace Transformers\cite{wolf-etal-2020-transformers} using the Adam optimizer\cite{kingma2014adam} at a learning rate of \(1.5\times10^{-5}\), cosine annealing, a batch size of 128, and a sequence length of 2048. To capture logit distributions, we set top-p=0.95, top-k=10, and temperature=2 for sampling during both training and inference. The training comprises two phases: (1) Inference Mode Fusion for 1 epoch (\(\lambda=0.1\)); (2) Inference-Training Co-fusion for 2 epochs (\(\lambda=0.5, \beta=0.5, w_1=0.1, w_2=1\)). For evaluation, we aggregate votes from three RewardBench\footnote{\url{https://huggingface.co/spaces/allenai/reward-bench}} reward models (Eurus-RM-7B \cite{eurus}, FsfairX-LLaMA3-RM-v0.1~\cite{dong2023raft,xiong2024iterative}, Starling-RM-7B-alpha\cite{starling2023}), resolving ties with the highest-performing model's score. The total computational cost is ~96 A100-80G GPU-hours.

% \vspace{-0.1cm}
\subsubsection{Baselines}
\label{baseline_intro}
We compare ProFuser against three categories of established baselines. 
(1) \textit{Original models}: Vicuna-7B-v1.5, Llama-2-7B-Chat, and MPT-7B-8K-Chat.
(2) \textit{Continual SFT}:  we utilize the Vicuna-7B-v1.5-CSFT as a baseline, which is subjected to continual SFT using the same dataset as ProFuser, ensuring a fair comparison.
(3) \textit{Model fusion}: this category features Vicuna-7B-v1.5-Fuse focusing on training mode fusion, Vicuna-7B-v1.5-SimulFuse performing simultaneous inference and training mode fusion, and Vicuna-7B-v1.5-ReverseFuse implementing training mode followed by inference mode fusion.
\vspace{-0.1cm}
\subsubsection{Evaluation}
We evaluate ProFuser across three dimensions.
(1) \textit{Knowledge}: the models' ability to understand and recall factual information is assessed using the MMLU dataset~\cite{hendrycks2020measuring}, which spans 57 diverse subjects such as elementary mathematics, US history, and other academic topics.
(2) \textit{Reasoning}: the models' general reasoning skills are appraised using challenging benchmarks such as HellaSwag~\cite{zellers2019hellaswag}, ARC-Challenge~\cite{clark2018think}, and WinoGrande~\cite{sakaguchi2021winogrande}. Additionally, mathematical reasoning is specifically assessed through the GSM8K.
(3) \textit{Safety}: the models' capability to generate outputs that align with factual correctness and common sense, relying on the TruthfulQA dataset~\cite{lin2021truthfulqa}.
Evaluations are conducted using the LM-Evaluation-Hardness framework~\cite{eval-harness}, following the standard metrics of the HuggingFace OpenLLM Leaderboard~\cite{open-llm-leaderboard}. For the GSM8K assessment, our approach follows the methodology outlined in Open-Instruct~\cite{wang2023far}.
\vspace{-0.1cm}
\subsection{Main Results}
\label{main_results}
Table \ref{tab:main-results} presents the performance of ProFuser compared to various baselines across six benchmarks, revealing several key insights:
Firstly, by integrating three source models into Vicuna-7B-v1.5-ProFuser using our proposed fusion methodology, we observe that ProFuser achieves the highest overall score among all evaluated methods. Specifically, it delivers a 3.09\% improvement over the baseline Vicuna-7B-v1.5, a substantial enhancement that is twice as large as the improvement achieved through continual supervised fine-tuning (Vicuna-7B-v1.5-CSFT). This result highlights ProFuser's ability to effectively leverage complementary strengths from multiple source models and surpass traditional fine-tuning approaches.

When comparing ProFuser against FuseLLM~\cite{wan2024knowledge}, Vicuna-7B-v1.5-ProFuser consistently outperforms FuseLLM across all benchmarks except GSM8K, where it still demonstrates a modest relative boost of 1.06\%. The slight underperformance on GSM8K can be attributed to challenges in mathematical reasoning tasks when fused source models exhibit frequent incorrect predictions during inference mode fusion. These inaccuracies introduce noise into the process and slightly limit its effectiveness.

Further comparison with alternative fusion strategies, including SimulFuse and ReverseFuse, shows that ProFuser consistently outperforms these approaches. Notably, ReverseFuse which prioritizes training mode fusion before inference mode, underperforms FuseLLM and even degrades overall performance. These findings validate the effectiveness of ProFuser’s sequential easy-to-hard fusion strategy: starting with simpler signals derived from inference mode (based on outputs generated by small source models) and gradually incorporating more complex knowledge-rich signals obtained during training mode (leveraging GPT-4 ground truth data). This progressive approach enables more effective integration of model strengths while avoiding early-stage overload or inefficiencies caused by premature exposure to high-complexity inputs.

\begin{figure}[htbp]
    \centering
    \includegraphics[width=\linewidth]{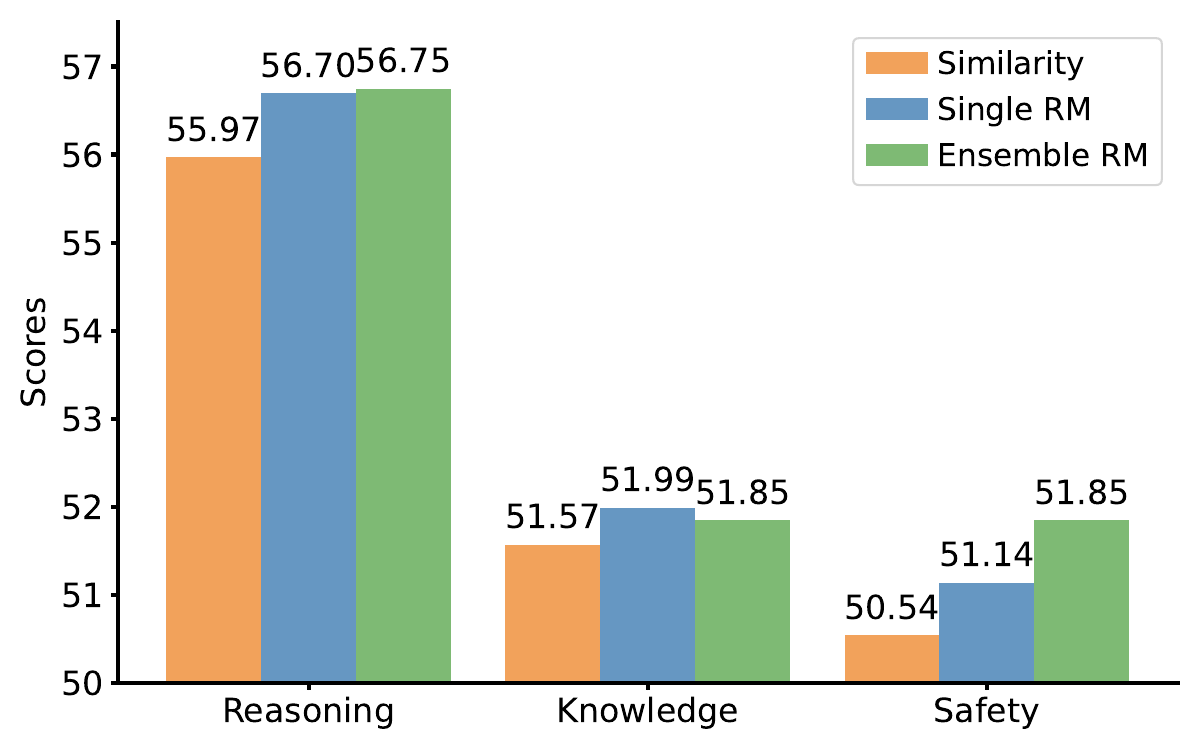}
    \caption{Results of different model advantage evaluation methods for the inference mode.}
    \label{advantage-measure}
    \vspace{-3mm}
\end{figure}

\subsection{Effect of Dual-Mode Advantage Evaluation}
\label{sec:advantage_evaluation}
To evaluate the effectiveness of dual-mode advantage evaluation, which combines inference and training modes, in capturing the strengths of source LLMs. While the Min-CE metric in training mode has been widely validated as a reliable measure for assessing model advantages~\cite{wan2024knowledge}, we focus on exploring inference-mode-based evaluation to complement it. Two experimental frameworks are employed: reference-based evaluation and reference-free evaluation.
(1) In the reference-based approach, we hypothesize that a source model is more advantageous if its output closely resembles GPT-4's responses. Similarity is measured across two dimensions—textual form using BLEU and ROUGE scores, and semantics using BERTScore\footnote{\url{https://huggingface.co/microsoft/deberta-v2-xlarge}} with their contributions combined into a weighted scoring formula:
\begin{equation}
\scalebox{0.84}{$\text{Score} = 0.25 \times \text{BLEU} + 0.25 \times \text{ROUGE} + 0.5 \times \text{BERTScore}$}.
\end{equation}
(2) The reference-free framework uses open-source reward models to directly score outputs without relying on GPT-4 references. For single reward models, the highest-scoring output is selected; when multiple reward models are used, majority voting determines the optimal response.

As shown in Figure~\ref{advantage-measure}, results reveal that reward model scoring consistently outperforms textual similarity metrics across benchmarks due to its ability to provide robust evaluations beyond surface-level text features. Textual similarity methods perform well for simple instructions with clear-cut answers but struggle with complex tasks requiring nuanced reasoning or detailed explanations where valid variations exist in phrasing or structure.
Moreover, integrating multiple reward models enhances performance on specific benchmarks such as TruthfulQA by leveraging their alignment with safety-related aspects while showing variability across other benchmarks due to domain-specific proficiencies among individual reward models.

\begin{table}[ht]
\resizebox{\linewidth}{!}{%
\begin{tabular}{lcccc} \hline
                       & \textbf{Knowledge}  & \textbf{Reasoning} & \textbf{Safety} & \textbf{Average} \\ \hline
MPT-7B-8K-Chat            & 41.55      &50.10 &  43.70 &  45.12    \\ 
Llama-2-7B-Chat        & 46.74 & 52.85 & 44.59    & 48.06    \\
Vicuna-7B-v1.5         & 51.17 & 53.92 & 50.37  & 51.82 \\ \hline
Vicuna-7B-v1.5-RMFuser    & 51.19 & 55.95  & 50.63  & 52.59  \\
Vicuna-7B-v1.5-GTLenFusr   & 51.30  & 56.15 & 50.80 & 52.75 \\
Vicuna-7B-v1.5-ProFuser & \textbf{51.85} & \textbf{56.75}  & \textbf{51.85} & \textbf{53.48}  \\ \hline
\end{tabular}%
}
\caption{Results of various progressive fusion strategies. The best-performing scores are \textbf{bolded}. For brevity, ground truth length-based fusion and reward model score-based fusion are abbreviated as GTLenFuser and RMFuser, respectively.}
\vspace{-0.2cm}
\label{fusion-strategy}
\end{table}

\begin{table*}[t]
\centering
\resizebox{0.9\textwidth}{!}{%
\begin{tabular}{lccccccc} \hline
                       & \textbf{MMLU}  & \textbf{HellaSwag} & \textbf{ARC}   & \textbf{Winogrande} & \textbf{GSM8K} & \textbf{TruthfulQA} & \textbf{Average} \\ \hline
MPT-7B-8K-Chat            & 41.55      & 77.52          & 46.93      & 71.35           & 11.00  & 43.70           & 48.68        \\ 
Llama-2-7B-Chat        & 46.74 & 78.63          & 52.90      & 71.74           & 16.40 & 44.59           &  51.83       \\
Vicuna-7B-v1.5         & 51.17 & 77.36     & 53.75 & 72.30      & 15.80 & 50.37      & 53.46   \\ \hline
\multicolumn{8}{c}{\textit{Separately Fusion}} \\ \hline
Vicuna-7B-v1.5-FuseMPT    & 50.95 &  \textbf{78.40}    &54.81  & \textbf{74.60}      & 17.44 & 50.37      & 54.43   \\
Vicuna-7B-v1.5-FuseLlama   & \textbf{51.44}  & 78.01     & \textbf{55.12} & 74.31      & \textbf{18.77} & \textbf{51.47}      & \textbf{54.85}   \\ \hline
\end{tabular}%
}
\caption{Results showing the influence of different source models on fusion performance. The best-performing scores are \textbf{bolded}. Vicuna-7B-v1.5 serves as the target model, fused separately with MPT-7B-8K-Chat and Llama-2-7B-Chat.}
\label{subsection_exp: src_model_contrib}
\vspace{0.1cm}
\end{table*}

\subsection{Effect of Progressive Fusion Strategy}
\label{sec:progressive_fusion_strategy}

To assess the effectiveness of ProFuser’s progressive fusion strategy, we conducted experiments comparing it against alternative approaches. ProFuser employs a stepwise integration process that begins with inference-mode fusion and transitions to training-mode fusion, guided by the intuition that ground truth (GT) data—typically more nuanced and knowledge-rich than source model outputs—should dictate the sequence of fusion. To further evaluate its robustness, we explored additional difficulty criteria for curriculum learning:
(1) \textit{Ground truth sequence length}: Instructions paired with longer responses were considered more challenging.
(2) \textit{Reward model score}: Lower scores on target model outputs indicated higher task difficulty.
The results are summarized in Table \ref{fusion-strategy}, leading to two key observations:

Firstly, ProFuser consistently outperformed all alternative strategies across various benchmarks, validating the effectiveness of its model-oriented progressive learning framework. By prioritizing inference-mode fusion first—leveraging simpler signals aligned closely with task-level outputs—it establishes a robust foundation before incorporating richer and more complex training-mode signals derived from GT data. This "easy-to-hard" paradigm is particularly effective because it enables gradual adaptation during knowledge transfer, allowing the fused model to integrate complementary strengths from both modes synergistically while minimizing potential conflicts or noise introduced by overly complex inputs early in the process.

Secondly, among alternative difficulty criteria tested: splitting tasks based on ground truth response length proved reliable as a measure of complexity. In contrast, using reward model scores as a criterion underperformed significantly. While reward-based metrics provide insights into task alignment at an output level (inference mode), they fail to capture deeper nuances associated with GT data required for comprehensive capability enhancement.

\subsection{Quantifying Source Model Contributions to Fusion}
\label{sec:impact_of_source_models}
To systematically analyze the influence of different source models on fusion effectiveness, we conducted comprehensive experiments combining Vicuna-7B-v1.5 (Vicuna) with two distinct source models: MPT-7B-8K-Chat (MPT) and Llama-2-7B-Chat (Llama). The experimental results, presented in Table \ref{subsection_exp: src_model_contrib}, reveal several significant findings:

\paragraph{\textit{Task-specific synergy with heterogeneous models.}}
The fusion with MPT demonstrates notably superior improvements on specific benchmarks, particularly HellaSwag and WinoGrande, compared to other evaluation metrics. A compelling observation emerges: despite MPT's inferior standalone performance relative to Llama on these benchmarks, its fusion benefits substantially exceed those achieved through Llama integration. This apparent paradox can be attributed to two primary factors: (1) While MPT exhibits lower overall performance compared to Llama, it demonstrates disproportionately strong capabilities specifically in HellaSwag and WinoGrande relative to its performance on other benchmarks, enabling it to contribute more specialized knowledge during the fusion process. (2) The architectural heterogeneity between MPT and Vicuna potentially facilitates the integration of complementary knowledge representations—a phenomenon less prevalent when fusing architecturally homogeneous models like Llama and Vicuna.

\paragraph{\textit{Stable generalization with homogeneous models.}}
Conversely, fusion with Llama yields consistent performance improvements across all benchmarks, exhibiting minimal task-specific variance. This stability can be primarily attributed to the homogeneity between Vicuna and Llama. The shared architectural characteristics minimize the divergence between source-target distributions during both training and inference phases, facilitating smooth information integration while maintaining high performance standards.

\begin{figure}[htbp]
    \centering
    \includegraphics[width=\linewidth]{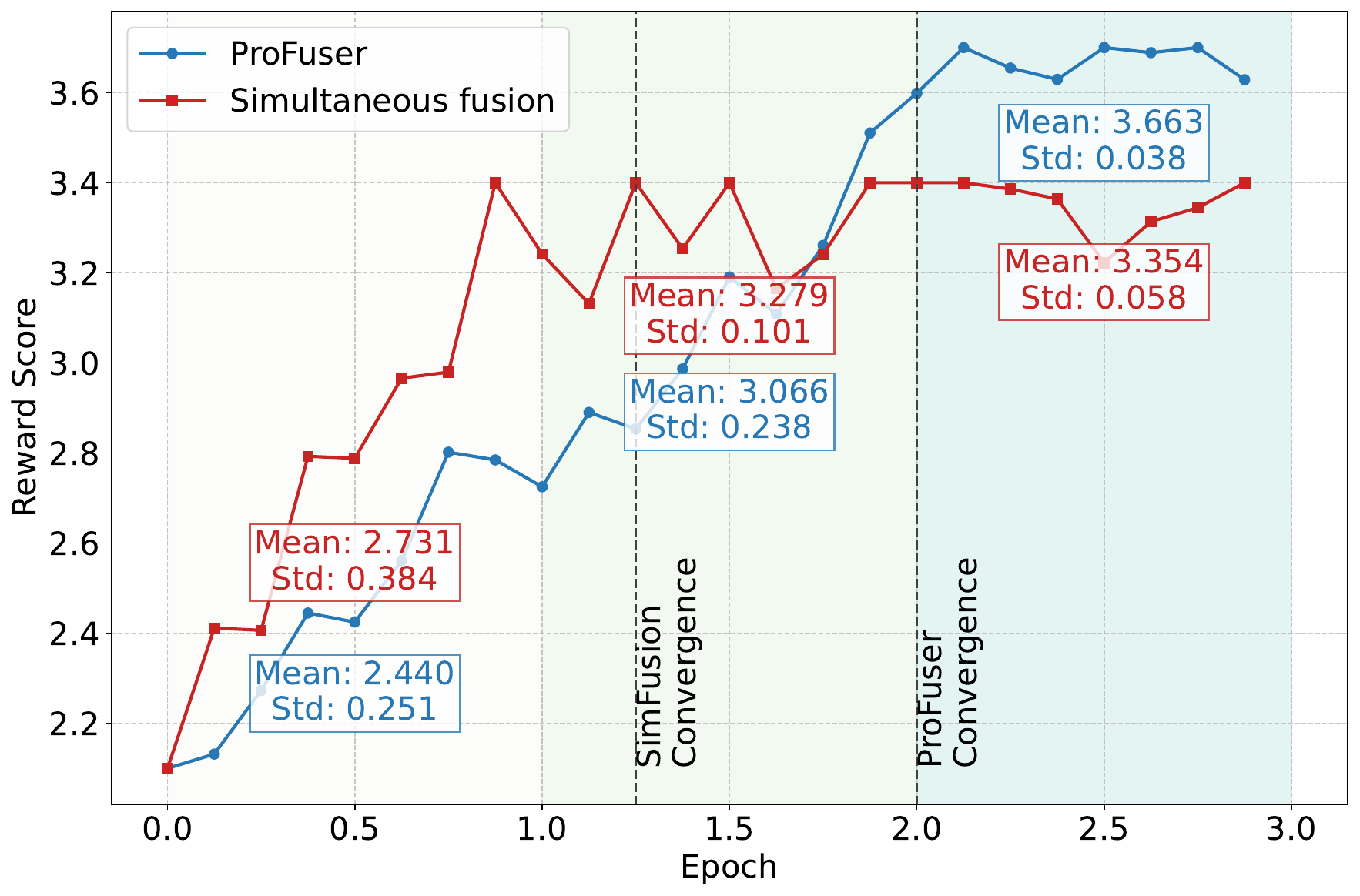}
    \caption{Training progress comparison between ProFuser and Simultaneous fusion approaches over three epochs. ProFuser (blue) achieves higher final reward model scores despite slower initial progress, converging at epoch 2. Simultaneous fusion (red) shows faster early improvement but converges earlier at epoch 1.25 with lower final performance. Mean and standard deviation of reward model scores are shown for each epoch.}
    \label{training_dynamic}
    \vspace{-3mm}
\end{figure}
% \vspace{-0.01cm}

\begin{figure}[htbp]
    \centering
    \includegraphics[width=0.9\linewidth]{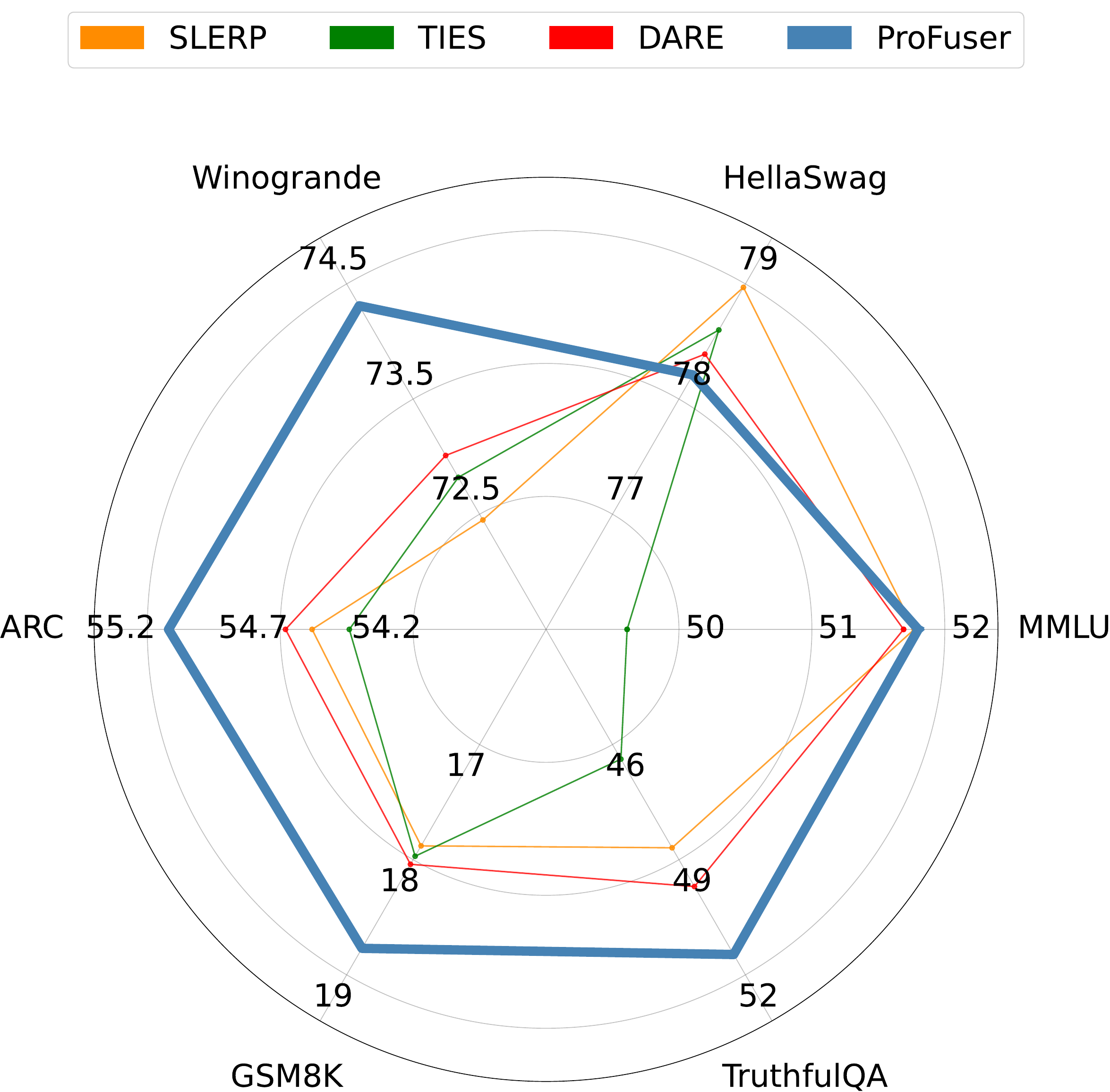}
    \caption{Comparison of ProFuser with three popular model merging methods (SLERP, TIES, and DARE) across six benchmarks.}
    \label{subsec-exp: compare with model merge}
    \vspace{-3mm}
\end{figure}
% \vspace{-0.03cm}

\subsection{Stability Analysis of Progressive Fusion Process}
To assess the impact of progressive fusion on the stability of the fusion process, we analyze the training dynamics of ProFuser in comparison to simultaneous fusion methods. Specifically, we randomly sample 2k data points from Orca-Best (excluding those used for constructing the training set) as an evaluation dataset. Throughout training, changes in reward model scores on this evaluation set are tracked using FsfairX-LLaMA3-RM-v0.1 as the scoring model.

As shown in Figure \ref{training_dynamic}, ProFuser exhibits a smoother and more consistent performance improvement during its fusion process, ultimately converging to a higher final point. In contrast, simultaneous fusion achieves faster initial gains but quickly stagnates at an early performance plateau.
This result underscores the advantages of ProFuser’s stepwise integration strategy: beginning with inference mode signals and gradually incorporating more complex knowledge-rich signals during training mode enables sustained improvements over time without prematurely exhausting learning potential. Conversely, simultaneous fusion combines all modes at once, resulting in rapid early progress but failing to fully leverage long-term optimization opportunities due to premature convergence.
These findings highlight that progressive fusion not only enhances stability throughout training but also achieves superior final outcomes by effectively balancing task complexity across different stages of learning.

% \vspace{-1mm}
\subsection{Comparison with Model Merging}
\label{sec:comparison_with_model_merging}
% \vspace{0.15cm}
To evaluate ProFuser's effectiveness in homogeneous model fusion scenarios, we conducted a comparison against various model merging methods. For fair comparison, we used Vicuna-7B-v1.5-CSFT and Llama-2-7B-Chat as baseline models in the merging experiments, ensuring alignment with ProFuser’s lightweight fine-tuning framework.  

As presented in Figure \ref{subsec-exp: compare with model merge}, ProFuser consistently achieves the highest scores across individual benchmarks such as MMLU, GSM8K, and TruthfulQA while also securing the highest overall average score.
While model merging methods exhibit competitive performance in specific cases, such as HellaSwag, where source models are comparably strong, their effectiveness diminishes when weaker models are incorporated. For instance, merging weaker models often leads to performance degradation of the base model, as observed in other benchmarks. In contrast, ProFuser maintains reliable fusion performance across diverse source model strengths, even with lightweight fine-tuning.  

These results highlight ProFuser’s robustness in heterogeneous fusion scenarios, where model merging approaches struggle to harmonize disparate capabilities. By dynamically balancing inference and training mode signals, ProFuser achieves more stable and generalizable improvements, underscoring its advantage over static merging techniques.

\section{Conclusion}
% \vspace{0.15}
Fusing the knowledge and capabilities of multiple LLMs can create stronger models more efficiently. We introduce ProFuser, a simple method that integrates the strengths of heterogeneous LLMs into a single LLM. Instead of relying solely on the training mode to capture the model's strengths in understanding ground truth, ProFuser also leverages the inference mode to capture the model's strengths in executing instructions, fully showcasing the model's advantages. Furthermore, ProFuser progressively learns from the inference mode to the training mode, based on the difference that ground truth (GPT-4 output) used in the training mode is more complex and detailed than the source LLM output in the inference mode, thus fully utilizing the advantages of both modes. Evaluated across six benchmarks and three dimensions, ProFuser performs significantly better than existing model fusion methods.

\section{Acknowledgments}
This work was supported by the National Natural Science Foundation of China (No. 62576368) and the Guangzhou Municipal Science and Technology Project (No. 2025B04J0018).

\bibliography{aaai2026}

\end{document}